\documentclass[10pt]{article}

\usepackage[letterpaper,top=0.55in,bottom=0.68in,left=1.50in,right=1.50in]{geometry}
\usepackage[T1]{fontenc}
\usepackage[utf8]{inputenc}
\usepackage{times}
\usepackage[round,authoryear]{natbib}

\usepackage{amsmath}
\usepackage{amssymb}
\usepackage{amsthm}
\usepackage{mathtools}

\usepackage{graphicx}
\usepackage{booktabs}
\usepackage{multirow}
\usepackage{pifont}
\usepackage{subcaption}
\usepackage{xcolor}
\usepackage{float}

\usepackage{algorithm}
\usepackage{algpseudocode}
\usepackage{tcolorbox}

\usepackage{microtype}
\usepackage{url}
\usepackage{xspace}

\usepackage{hyperref}
\hypersetup{
  colorlinks=true,
  linkcolor=blue!60!black,
  citecolor=blue!60!black,
  urlcolor=blue!60!black,
  pdftitle={LIBERO-Dynamic: Benchmarking Robotic Manipulation in Dynamic Environments},
  pdfauthor={Shaojie Ren, Jingwen Sun, Zezhi Liu, Hongjin Chen, Xin Jin, Wei Zou, and Zhibo Chen}
}
\usepackage[capitalize,noabbrev]{cleveref}

\newcommand{\bench}{\textsc{LIBERO}-Dynamic\xspace}

\newcommand{\libero}{\textsc{LIBERO}\xspace}

\theoremstyle{definition}

\newcommand{\parhead}[1]{\paragraph{#1}}

\definecolor{findingbluetitle}{HTML}{EAF2FA}
\definecolor{findingblueframe}{HTML}{7893B0}
\newtcolorbox{findingbox}[1][]{%
  colback=white,
  colbacktitle=findingbluetitle,
  colframe=findingblueframe,
  coltitle=black,
  boxrule=0.4pt,
  arc=1.5pt,
  left=6pt, right=6pt, top=4pt, bottom=4pt,
  fonttitle=\bfseries,
  title={#1},
}
\newcounter{finding}
\newcommand{\finding}[2]{%
  \refstepcounter{finding}%
  \begin{findingbox}[Finding \thefinding: #1]
    #2
  \end{findingbox}
}

\newcommand{\Lonefam}{linear ping-pong\xspace}
\newcommand{\Ltwofam}{Catmull-Rom spline\xspace}
\newcommand{\Lthreefam}{multi-waypoint zigzag\xspace}

\title{LIBERO-Dynamic: Benchmarking Robotic Manipulation in Dynamic Environments}
\author{Shaojie Ren, Jingwen Sun, Zezhi Liu, Hongjin Chen, Xin Jin, Wei Zou, and Zhibo Chen}
\date{}

\begin{document}
\raggedbottom

\begin{center}
{\LARGE\bfseries LIBERO-Dynamic: Benchmarking Robotic\\
Manipulation in Dynamic Environments\par}
\vspace{0.8em}
{\normalsize
Shaojie Ren\textsuperscript{1,2,3}\quad
Jingwen Sun\textsuperscript{4}\quad
Zezhi Liu\textsuperscript{5}\quad
Hongjin Chen\textsuperscript{6}\\[-0.05em]
Xin Jin\textsuperscript{7}\quad
Wei Zou\textsuperscript{3,\ensuremath{\dagger}}\quad
Zhibo Chen\textsuperscript{4,\ensuremath{\dagger}}\par}
\vspace{0.6em}
{\small
\textsuperscript{1}University of Chinese Academy of Sciences\quad
\textsuperscript{2}Beijing Zhongguancun Academy\\[-0.08em]
\textsuperscript{3}Institute of Automation, Chinese Academy of Sciences\\[-0.08em]
\textsuperscript{4}University of Science and Technology of China\quad
\textsuperscript{5}Nankai University\\[-0.08em]
\textsuperscript{6}Harbin Institute of Technology\quad
\textsuperscript{7}Eastern Institute of Technology, Ningbo\\[0.25em]
\textsuperscript{\ensuremath{\dagger}}Corresponding authors.\\[0.25em]
\href{https://renshaojie233.github.io/LIBERO-Dynamic/}{Project page: renshaojie233.github.io/LIBERO-Dynamic}
\par}
\end{center}
\vspace{0.25em}

\begin{abstract}
Most simulation benchmarks for robotic manipulation keep target objects static and treat policy inference as instantaneous. These assumptions simplify repeatable evaluation, but they hide how a moving target evolves during computation and how stale an observation may be when its predicted action is executed. We introduce \bench, a controlled dynamic extension of all 40 \libero tasks. It drives objects through contact friction along three trajectory families and three speed tiers, while continuously advancing the physical world according to calibrated, model-specific inference latency. The same motion mechanism also supports training: we replay verified static demonstrations and compensate robot actions online using the target's displacement relative to the reference, producing 9{,}000 quality-filtered dynamic trajectories. We evaluate six pretrained policies on a shared set of 1{,}800 dynamic conditions per configuration, compare execution strategies and action horizons with the checkpoint fixed, and fine-tune three architectures using dynamic and mixed data. The experiments show that faster motion degrades every evaluated policy, but not uniformly; execution timing and horizon can change the behavior of the same policy; and, in the measured sweep, asynchronous replanning benefits from frequent feedback only after the action queue can be sustained without underflow. Dynamic demonstrations improve each evaluated architecture under at least one training recipe, while the magnitude of improvement varies with temporal conditioning, data composition, and training budget. In a small-scale physical study using policies trained on real-robot dynamic demonstrations, we observe execution-strategy differences consistent with those found in simulation, with task-dependent effects, providing complementary evidence for the corresponding benchmark findings. By making object motion, inference latency, and action execution explicit and controllable within a broad task suite, \bench provides a reproducible platform for evaluating and improving robot policies for moving-target manipulation.

\end{abstract}
\vspace{1.0em}

\section{Introduction}
\label{sec:intro}

\begin{figure}[!ht]
  \centering
  \includegraphics[width=\linewidth]{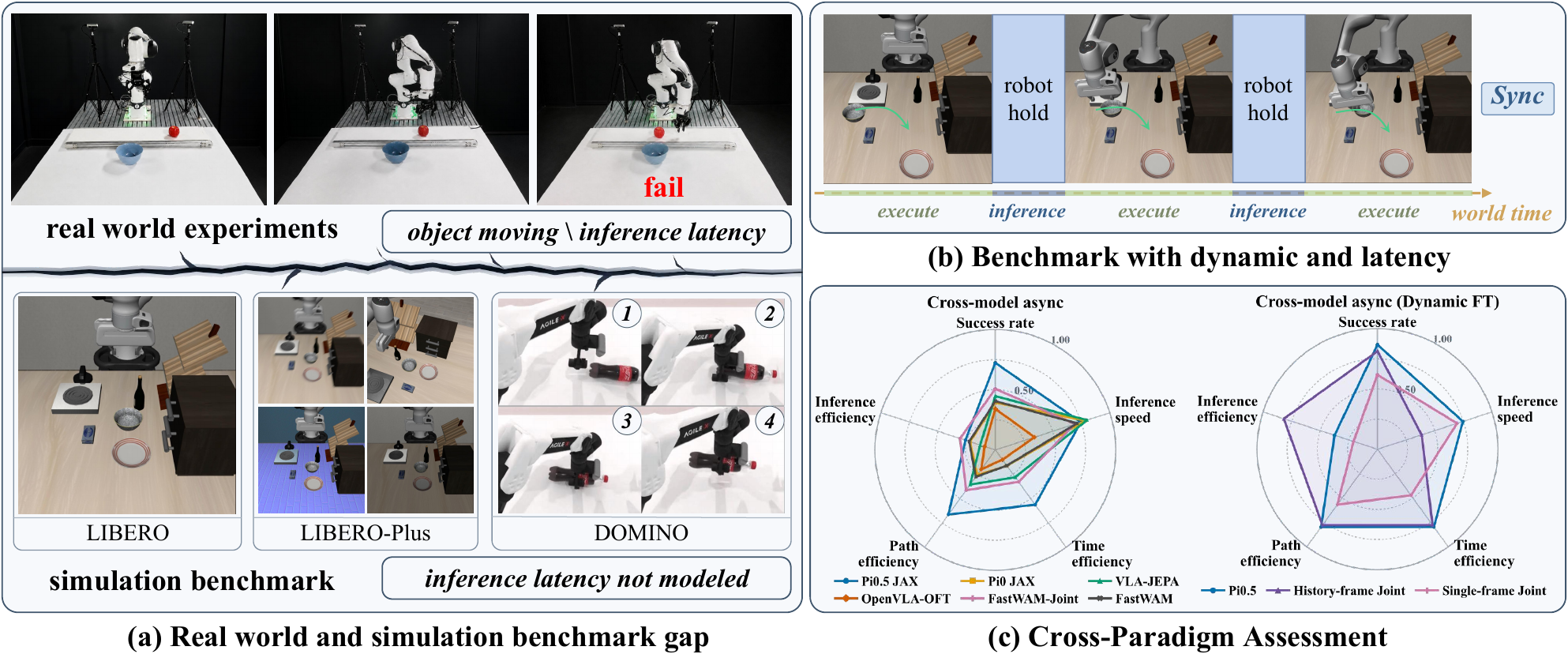}
  \caption{Overview of \bench: \textbf{(a)} dynamic real-world manipulation and conventional simulation evaluation; \textbf{(b)} latency-aware evaluation; and \textbf{(c)} comparisons across execution and training settings.}
  \label{fig:overview}
\end{figure}

General-purpose robot policies---including vision-language-action (VLA) models and policies built on video or world models---have advanced rapidly with growing data scale, architectural innovation, and improved post-training techniques~\citep{brohan2023rt2,octo2024,kim2024openvla,black2024pi0,yuan2026fastwam}.
Reliable comparison increasingly depends on benchmarks that evaluate not only task completion, but also behavior under deployment-relevant conditions.

Real-robot platforms such as RoboArena~\citep{atreya2025roboarena}, RoboChallenge~\citep{yakefu2025robochallenge}, and RoboDojo~\citep{chen2026robodojo} provide direct evidence under physical deployment, whereas simulation benchmarks such as \libero~\citep{liu2023libero}, RLBench~\citep{james2020rlbench}, and RoboTwin~\citep{mu2024robotwin} enable repeatable evaluation across broad task collections. Recent extensions including LIBERO-Plus~\citep{fei2025liberoplus}, LIBERO-PRO~\citep{zhou2025liberopro}, and LIBERO-X~\citep{liberox2026} further examine robustness and generalization under controlled variations.

Dynamic manipulation adds a critical temporal dimension. In common step-based simulation evaluations, policy inference is typically treated as instantaneous, or the physical world does not continue to advance during inference, leaving observations and actions ideally aligned in time. In a static scene, waiting for inference does not change the state of the target object; when the object continues to move, however, the observation underlying a returned action may already be stale. DOMINO~\citep{fang2026domino} studies moving-object interaction, while DOM~\citep{xie2026dynamicvla} and ReflexBench~\citep{reflex2026} also account for inference latency and compare execution mechanisms. These advances motivate a complementary question: how do object motion, execution timing, and training choices affect policy performance when the underlying manipulation tasks are held consistent?

To address this gap, we introduce \bench, a dynamic manipulation benchmark that combines controlled object motion with model-specific inference latency in a unified simulation loop; its overall framework is shown in \cref{fig:benchmark_overview}.
Across \libero's 40 tasks, \bench introduces three trajectory families and three motion-speed tiers, using a friction-driven invisible contact platform to preserve contact interactions among the robot, target, and surrounding objects.
The physical world continues to evolve during policy inference, and the benchmark advances simulation according to pre-calibrated model latency so that observation--action staleness can be evaluated explicitly. With the checkpoint fixed, we conduct controlled comparisons of synchronous execution, the LeRobot asynchronous queue~\citep{cadene2026lerobot}, and Real-Time Chunking~(RTC)~\citep{black2025rtc}, while treating VLASH~\citep{vlash2025}, which uses a separate checkpoint, as a future-state-aware system reference.

Under a unified protocol, we evaluate diverse robot policies and compare dynamic manipulation performance across execution mechanisms and action horizons.
We further conduct controlled fine-tuning experiments on three model families using dynamic demonstrations and mixed static--dynamic data, examining how gains from dynamic data vary with the form of temporal conditioning.
A compact real-robot study, using policies trained on physical dynamic demonstrations in a DROID-style setup~\citep{khazatsky2024droid}, provides supplementary evidence for the execution-timing observations under physical sensing and actuation.

In summary, this work makes the following contributions:
\begin{enumerate}
    \item We introduce \bench, a dynamic robotic manipulation benchmark that combines controlled object motion with model-specific inference latency. It defines a controlled spatiotemporal evaluation space through motion pattern, speed, trajectory phase, and perception--action delay for comparing dynamic responsiveness across robot policies.
    \item We introduce a method for converting successful static demonstrations into dynamic ones: invisible conveyors drive objects through contact friction, while static manipulation references are adapted to changing target states through online motion compensation. Using this method, we construct a dataset of 9{,}000 dynamic trajectories spanning multiple motion patterns, speeds, spatial trajectories, and motion phases.
    \item We evaluate representative models and execution strategies under a unified protocol and use controlled fine-tuning to analyze the relationship among dynamic data, temporal conditioning, and execution. Dynamic data improves several evaluated policies, while the gains vary with the form of temporal conditioning, providing empirical evidence for model training and execution design in dynamic manipulation.
\end{enumerate}

\section{Related Work}
\label{sec:related}

\parhead{Robot Policies and Temporal Modeling.}
General-purpose robot policies span autoregressive action-token prediction, continuous action generation, and video- or world-model-based control, represented by RT-2 and OpenVLA, OpenVLA-OFT and the $\pi$ family, and VLA-JEPA and Fast-WAM, respectively~\citep{brohan2023rt2,kim2024openvla,kim2025openvlaoft,black2024pi0,pi052025,sun2026vlajepa,yuan2026fastwam}.
Recent methods for dynamic settings further incorporate observation history, future-state prediction, or continuous inference--execution coordination into policy design~\citep{fang2026domino,reflex2026,xie2026dynamicvla}.
Because these systems also differ in training data and execution protocol, direct comparison across reported results is difficult.
We evaluate representative policy families under a common task distribution and timing protocol, and study how temporal conditioning interacts with dynamic-data fine-tuning.

\parhead{Benchmarks and Data for Dynamic Manipulation.}
LIBERO, RLBench, and RoboTwin support reproducible multi-task manipulation evaluation, while LIBERO-Plus, LIBERO-PRO, LIBERO-X, VLA-Arena, and LIBERO-Safety extend this setting toward robustness, generalization, or safety under controlled variations~\citep{liu2023libero,james2020rlbench,mu2024robotwin,fei2025liberoplus,zhou2025liberopro,liberox2026,zhang2026vlaarena,cui2026liberosafety}.
More recent benchmarks study within-rollout dynamics: DOMINO provides 35 tasks with hierarchical motion complexity and large-scale dynamic demonstrations; DynamicVLA introduces DOM for dynamic interaction, perception, and generalization in simulation; and ReflexBench couples six reaction-critical tasks with configurable latency under synchronous and asynchronous execution~\citep{fang2026domino,xie2026dynamicvla,reflex2026}.
\Cref{tab:benchmark_comparison} summarizes the benchmark properties most directly related to our scope.

\begin{table}[H]
  \centering
  \caption{Comparison of simulation benchmarks. Task counts exclude perturbation variants; ``--'' indicates that DOM v1 reports nine evaluation dimensions rather than a comparable semantic-task count. Controlled dynamics denotes prescribed within-rollout motion; data mixture compares dynamic-only and mixed static--dynamic training. Checks indicate reported features or analyses; crosses indicate not reported.}
  \label{tab:benchmark_comparison}
  \newcommand{\benchyes}{\textcolor{green!55!black}{\ding{51}}}
  \newcommand{\benchno}{\textcolor{red!75!black}{\ding{55}}}
  \renewcommand{\arraystretch}{1.12}
  \footnotesize
  \begin{tabular*}{\textwidth}{@{\extracolsep{\fill}}lcccccc@{}}
    \toprule
    Benchmark & \shortstack{Sim.\\tasks} & \shortstack{Controlled\\dynamics} & \shortstack{Latency-\\aware} & \shortstack{Execution\\comparison} & \shortstack{Dynamic\\demos} & \shortstack{Data\\mixture} \\
    \midrule
    LIBERO-Plus    & 40     & \benchno  & \benchno  & \benchno  & \benchno  & \benchno  \\
    LIBERO-X       & 600    & \benchno  & \benchno  & \benchno  & \benchno  & \benchno  \\
    LIBERO-Safety  & 75     & \benchyes & \benchno  & \benchno  & \benchyes & \benchno  \\
    DOMINO         & 35     & \benchyes & \benchno  & \benchno  & \benchyes & \benchyes \\
    DynamicVLA/DOM & --     & \benchyes & \benchyes & \benchyes & \benchyes & \benchno  \\
    ReflexBench    & 6      & \benchyes & \benchyes & \benchyes & \benchyes & \benchno  \\
    \midrule
    \textbf{LIBERO-Dynamic} & \textbf{40} & \benchyes & \benchyes & \benchyes & \benchyes & \benchyes \\
    \bottomrule
  \end{tabular*}
\end{table}

These benchmarks address complementary gaps. DOMINO studies motion complexity and static--dynamic co-training; DOM evaluates dynamic interaction, perception, and generalization, and analyzes continuous inference and latency-aware action streaming. ReflexBench compares synchronous and asynchronous execution under configurable latency. \bench retains the goals of all 40 LIBERO tasks and applies matched trajectory and speed controls with pre-calibrated model latency. This setting supports action-horizon sweeps and queue-underflow analysis, together with comparisons of dynamic-only and mixed fine-tuning across policy structures. Its emphasis is on how motion, execution timing, and training choices affect performance within a shared task distribution.

\parhead{Latency-Aware Execution of Action-Chunking Policies.}
Action chunking trades feedback frequency for temporal consistency: long execution horizons reduce policy queries, whereas short horizons refresh observations more often but can exhaust the action queue under non-negligible inference latency.
LeRobot implements a producer--consumer asynchronous queue that overlaps action prediction and execution~\citep{cadene2026lerobot}.
RTC conditions generation on committed actions through inference-time inpainting, while training-time RTC learns prefix-conditioned continuation during training to avoid the additional guidance pass~\citep{black2025rtc,black2025trainingrtc}.
A2C2 adds a lightweight per-step correction head driven by the latest observation, and VLASH aligns prediction with execution by rolling the robot state forward over the expected delay~\citep{sendai2025a2c2,vlash2025}.
FASTER reduces time to first action through horizon-aware flow-sampling schedules~\citep{lu2026faster}.
Together, these works address latency and action-chunk continuity at the levels of execution, training, online correction, and sampling.

\section{The \bench Benchmark}
\label{sec:benchmark}

\begin{figure}[t]
  \centering
  \includegraphics[width=\linewidth]{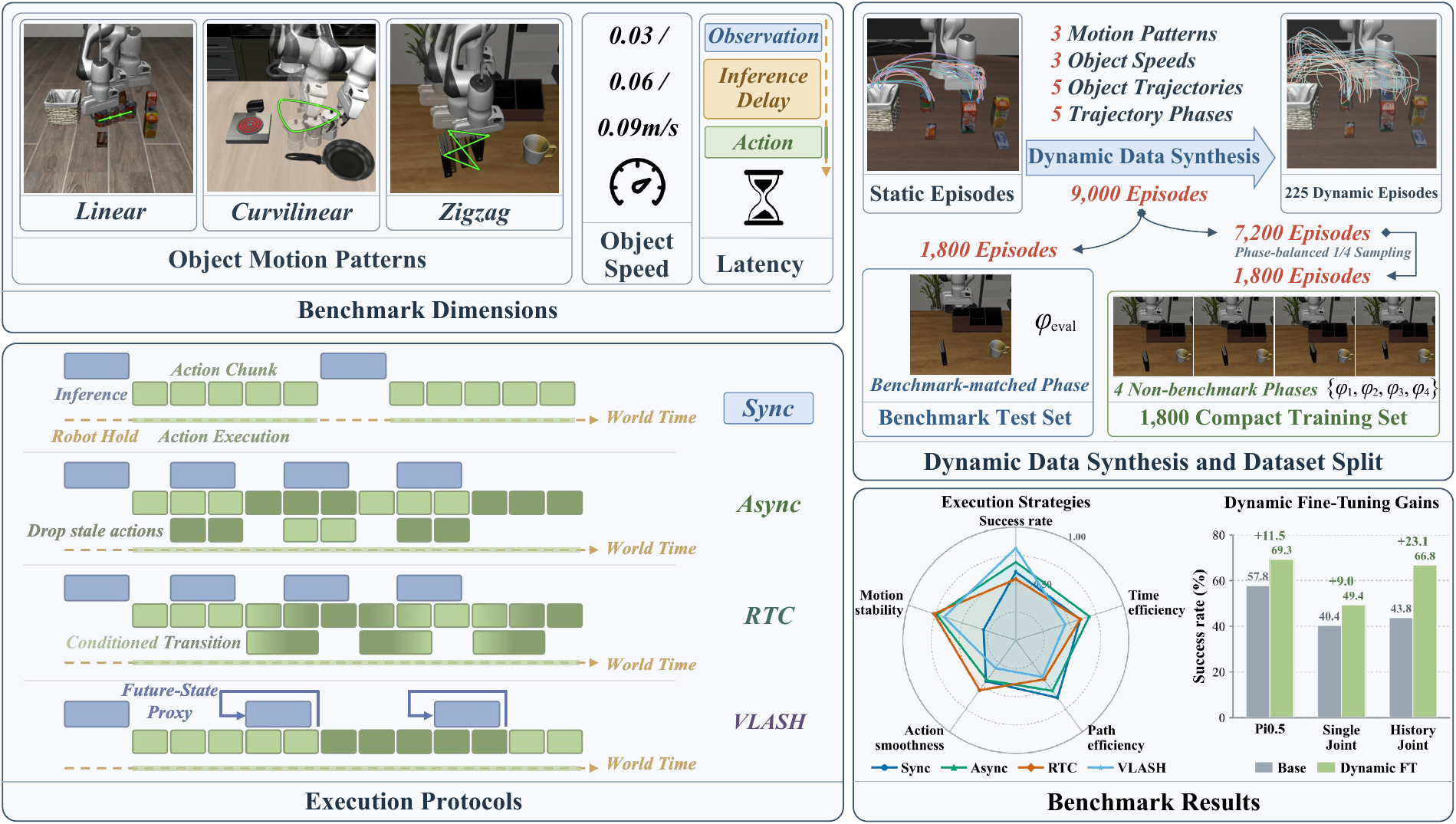}
  \caption{Benchmark dimensions, dynamic-data construction, execution protocols, and representative results. Blue denotes inference; alternating light and dark green denote actions from consecutive predictions.}
  \label{fig:benchmark_overview}
\end{figure}

\bench studies manipulation in which task-relevant objects move continuously while policy inference advances the physical world according to calibrated model latency. The action returned by a policy may therefore respond to an earlier target state. We instantiate this setting on all 40 \libero tasks, with 1{,}800 fixed episodes per model--execution configuration (\cref{fig:benchmark_overview}).

\subsection{Tasks, Motion Families, and Evaluation Scale}
\label{sec:bench:tasks}
\label{sec:bench:motion}

Each of the four ten-task \libero suites contributes three motion families and five valid trajectory variants per family, yielding $40\times3\times5=600$ dynamic configurations. The families capture constant-speed linear reciprocation, a smooth closed Catmull--Rom path~\citep{catmull1974splines}, and a piecewise-linear multi-waypoint zigzag. They differ in direction continuity and turning structure; their labels do not prescribe a difficulty order. Additional target and trajectory statistics are in \cref{app:benchmark_details}.

Every configuration is evaluated at 0.03, 0.06, and 0.09\,m/s, producing $600\times3=1{,}800$ episodes per deployment configuration. All methods share the same initial states, paths, motion phases, and success conditions.

\subsection{Contact-Driven Motion and Continuous Physics}
\label{sec:bench:sim}

Instead of overwriting object poses, \bench places an invisible rigid platform beneath each target and transfers its commanded motion through MuJoCo contact friction~\citep{todorov2012mujoco}. The target remains force-driven, can separate from the platform when grasped, and interacts with the robot and scene through the standard solver.

Control runs at $f_c=20$\,Hz. Measured inference latency $\tau$ is quantized to $D=\lceil\tau f_c\rceil$ control steps, and simulation advances by $D/f_c$ before the returned action becomes available. During this interval, physics and object motion continue while the robot either holds or executes queued actions according to the selected protocol. All strategies use the same latency quantization and physics advancement rule; contact and calibration details are reported in \cref{app:benchmark_details,app:execution_protocol}.

\subsection{Static-to-Dynamic Demonstration Conversion}
\label{sec:bench:data}

\paragraph{Reference construction.}
For each task, we replay candidates from the official static teleoperation data with object motion disabled. The primary template is the replay-valid candidate that completes the task earliest within a deterministic scan. Replaying this template records end-effector, gripper, target, and destination trajectories; gripper state and target proximity identify grasp, release, and fixture-contact phases.

\paragraph{Dynamic replay and validation.}
At each step, the generator maps the reference target--end-effector relation into the target's current frame and adds a smoothed, bounded translational correction to the reference action. Compensation is activated around the relevant grasp, placement, or fixture-contact phase and falls back to spatial proximity when a temporal phase is unavailable. All robot commands pass through \texttt{env.step(action)}, and target motion remains contact-driven. An episode is retained only after task success, measured dynamic motion, and physical diagnostics all pass. A failed primary rollout enters a finite, deterministic fallback sequence that may change the verified reference, controller route, or grasp/place margins; every attempt is recorded, and failure of the full sequence excludes the condition. Exact equations, thresholds, and retry logic are in \cref{app:data_generation}.

\begin{figure*}[t]
  \centering
  \includegraphics[width=0.96\textwidth]{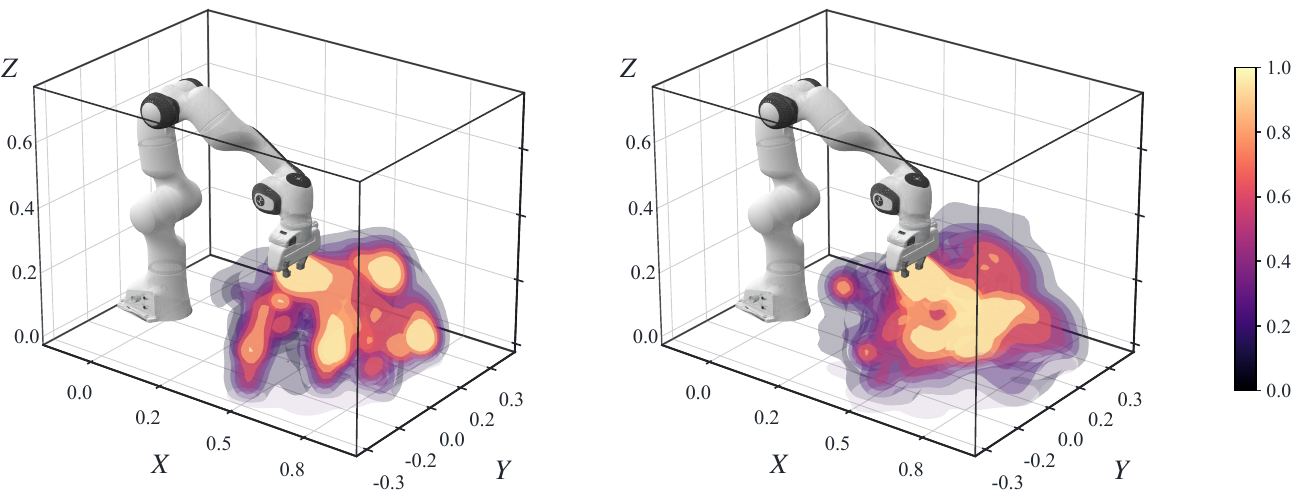}
  \caption{Normalized end-effector position density in the robot base frame for original \libero demonstrations (left) and generated \bench demonstrations (right).}
  \label{fig:eef_coverage}
\end{figure*}

For every one of the 1{,}800 task--trajectory--speed conditions, generation covers five initial motion phases, producing 9{,}000 successful dynamic demonstrations. Formal evaluation fixes one phase by advancing the scene for 40 control steps before the first observation. The remaining four phases form 7{,}200 training candidates, from which the compact split selects 1{,}800 trajectories. Training and evaluation have no episode-ID overlap but share tasks, trajectory families, paths, and speeds; the study therefore measures phase-separated, condition-matched adaptation rather than generalization to unseen tasks or trajectories. \Cref{fig:eef_coverage} visualizes the resulting end-effector coverage.

\subsection{Latency-Aware Execution Protocols}
\label{sec:bench:protocol}

A policy predicts $H$ actions, while the effective horizon $K\leq H$ limits how many may be adopted before replanning. We evaluate four protocols. \textbf{Synchronous hold} waits after the current chunk is exhausted. \textbf{Asynchronous queue} overlaps inference and action execution and blends overlapping predictions. \textbf{RTC} conditions continuation on committed actions and expected delay. \textbf{VLASH} rolls the robot state forward over the delay and hands off a returned chunk without weighted blending. Trigger and queue-update details are in \cref{app:execution_protocol}.

Latency is the end-to-end wall-clock time from observation submission to chunk receipt. Each deployment path is measured after warm-up on the target GPU and its median is frozen. Protocols share all benchmark conditions and physics parameters, while protocol-specific inputs and latency remain part of each complete configuration; comparisons therefore do not isolate a single mechanism causally.

\subsection{Evaluation Metrics}
\label{sec:bench:metrics}

An episode succeeds when the original \libero BDDL goal is satisfied within 600 control steps. The primary metric is task success rate (SR), reported overall and at each speed. We use four descriptive quantities. Tables report frozen latency $\tau$. For successful condition $(s,i)$, let $c_T$ be the reference-to-current step-count ratio, capped at 1; time efficiency is $E_T=\sqrt{p\,\bar c_T}$, where $p$ is SR. Replacing steps with end-effector path length gives path efficiency $E_P=\sqrt{p\,\bar c_P}$. Query efficiency is $E_I=1000N_{\mathrm{succ}}/N_{\mathrm{query}}$.

$E_T$ and $E_P$ combine success coverage with efficiency conditional on success and are comparable only within the same declared reference set; synchronous waiting time is excluded from the control-step count. Radar mappings and display ranges are in \cref{app:metric_details}. We also report \textbf{steady underflow}: action-queue-empty steps while inference is pending, after at least one predicted action has been executed, per 1{,}000 executed steps. It is used only to diagnose asynchronous queue continuity.

\section{Experiments}
\label{sec:experiments}

\parhead{Experimental Setup.}\label{sec:exp:setup} We evaluate six released \libero policies: flow-matching $\pi_{0.5}$ and $\pi_0$~\citep{pi052025,black2024pi0}, latent-world-model VLA-JEPA~\citep{sun2026vlajepa}, OpenVLA-OFT~\citep{kim2025openvlaoft}, and video-model-based FastWAM and FastWAM-Joint~\citep{yuan2026fastwam}. Cross-model evaluation uses native checkpoints and interfaces without dynamic training; the training study additionally includes a five-frame history variant of FastWAM-Joint. Every configuration uses the same 1{,}800 benchmark episodes and success criteria. Deployment-path latency is calibrated and frozen on an RTX~5090.

\subsection{Cross-Model Evaluation under Dynamics}
\label{sec:exp:crossmodel}

All six policies use asynchronous execution. For each model, $K^*$ is the highest-SR point among measured $K{=}1$, even horizons within its native action length, and the native odd endpoint when applicable. It is an oracle best-measured point on the fixed evaluation population, rather than a validation-selected setting. \Cref{tab:crossmodel} reports results at $K^*$; complete sweeps and motion-family results are in \cref{fig:ksweep,tab:motion_breakdown}.

\begin{table}[H]\centering
\caption{Six pretrained policies under asynchronous execution at their best measured $K^*$ on the fixed 1{,}800 episodes. Ret. is dynamic All SR divided by the speed $=0$ result.}
\label{tab:crossmodel}\scriptsize\setlength{\tabcolsep}{1.4pt}\begin{tabular*}{\linewidth}{@{\extracolsep{\fill}}lccccccccccc@{}}\toprule
 & $\tau$ & & \multicolumn{2}{c}{Static baseline} & \multicolumn{4}{c}{Dynamic SR (\%)} & & & \\
\cmidrule(lr){4-5}\cmidrule(lr){6-9}
Model & (ms) & $D$ & speed $=0$ & Ret. (\%) & 0.03 & 0.06 & 0.09 & All & $E_T$ & $E_P$ & $E_I$ \\
\midrule
$\pi_{0.5}$ (JAX) & 60 & 2 & 96.9 & \textbf{59.6} & 64.8 & 57.5 & 51.2 & \textbf{57.8} & \textbf{.715} & \textbf{.727} & 3.14 \\
FastWAM-Joint & 64 & 2 & 98.5 & 41.2 & 49.3 & 39.5 & 33.0 & 40.6 & .593 & .606 & \textbf{3.70} \\
VLA-JEPA & 56 & 2 & 97.2 & 36.7 & 41.0 & 37.5 & 28.7 & 35.7 & .566 & .577 & 1.59 \\
FastWAM & 63 & 2 & 97.6 & 32.9 & 34.2 & 32.5 & 29.5 & 32.1 & .503 & .533 & 2.68 \\
$\pi_0$ (JAX) & 57 & 2 & 94.2 & 33.4 & 36.2 & 32.7 & 25.7 & 31.5 & .507 & .530 & 2.65 \\
OpenVLA-OFT & 93 & 2 & 97.1 & 28.1 & 34.3 & 26.8 & 20.7 & 27.3 & .479 & .504 & 1.12 \\
\bottomrule
\end{tabular*}
\end{table}

Overall dynamic SR ranges from 27.3\% to 57.8\%. Normalization by speed $=0$ largely preserves the ordering but swaps FastWAM and $\pi_0$, so static ability alone does not determine the dynamic ranking. Raising object speed from 0.03 to 0.09\,m/s lowers every model by 4.7--16.3 points, with model-dependent declines.

\finding{Faster target motion reduces success for every evaluated model, but speed sensitivity differs across models.}{%
All six models decline as motion accelerates, but by different amounts and with changes in relative ordering.
}

\subsection{Execution Strategy and Action Horizon}
\label{sec:exp:strategy}

\begin{figure}[t]
  \centering
  \includegraphics[width=\linewidth]{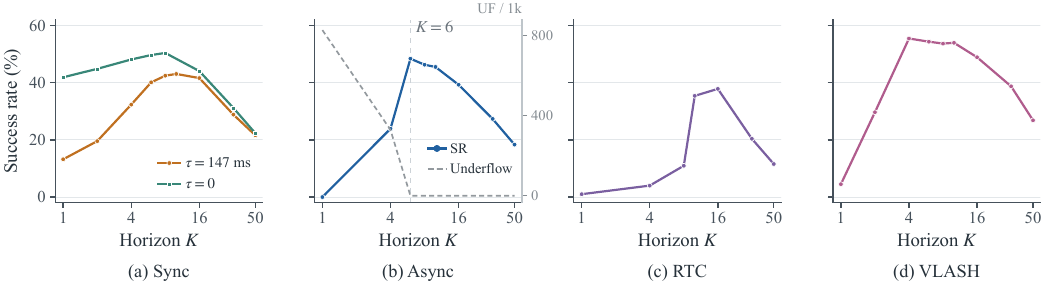}
  \caption{$\pi_{0.5}$ (PyTorch) SR versus action horizon under \textbf{(a)} Sync, \textbf{(b)} Async, \textbf{(c)} RTC, and \textbf{(d)} VLASH. Panel (a) includes a zero-latency diagnostic; panel (b) overlays steady underflow (right axis).}
  \label{fig:strategy_ksweep}
\end{figure}

We fix the $\pi_{0.5}$ (PyTorch) checkpoint and compare Sync, Async, and RTC (fixed $=10$). Sync and Async use the 147\,ms standard path; RTC is separately calibrated at 170\,ms. VLASH uses its own future-state-aware checkpoint and native interface~\citep{vlash2025}, so it is reported as a system reference rather than a same-checkpoint ablation. \Cref{fig:strategy_ksweep} shows the four sweeps; extended RTC settings and exact best-point metrics are in \cref{app:rtc,tab:strategy}.

With synchronous hold, every replanning event introduces an approximately 147\,ms pause. The zero-latency gap therefore grows from 2.4 points at $K{=}16$ to 25.3 points at $K{=}2$: more frequent replanning also creates more frequent stalls. At their best measured horizons, Async, Sync, and RTC obtain 48.4\%, 43.1\%, and 37.8\% SR with the same checkpoint. Async also uses the fewest completion steps on common successes, whereas RTC produces the lowest action variation and end-effector acceleration (\cref{tab:strategy_behavior}). RTC's extra conditioning increases latency to 170\,ms and changes queue updates, so its lower SR cannot be attributed to latency alone. VLASH reaches 55.4\% as a separately trained complete system.

The Async sweep separates two regimes. At $K{=}1$ and 4, steady underflow is 828.3 and 329.5 per 1{,}000 steps and SR is 0.0\% and 23.9\%; at $K{=}6$, underflow reaches zero and SR peaks at 48.4\%. For measured $K{\geq}6$, underflow remains zero while SR declines to 39.3\% at $K{=}16$. In this fixed-checkpoint sweep, queue continuity sets the minimum underflow-free horizon; beyond it, shorter open-loop execution is associated with higher success.

\finding{Execution timing changes dynamic-task performance even when model weights are fixed.}{%
Waiting, queuing, and plan handoff alter the world state to which the same checkpoint ultimately responds.
}

\finding{Once queue underflow is avoided, more frequent feedback can improve success.}{%
In the fixed-checkpoint $\pi_{0.5}$ asynchronous sweep, shorter horizons perform better within the measured underflow-free regime; the preferred horizon remains model- and recipe-dependent.
}

\subsection{Training with Dynamic Data}
\label{sec:exp:training}

\begin{figure*}[t]
\centering
\includegraphics[width=\textwidth]{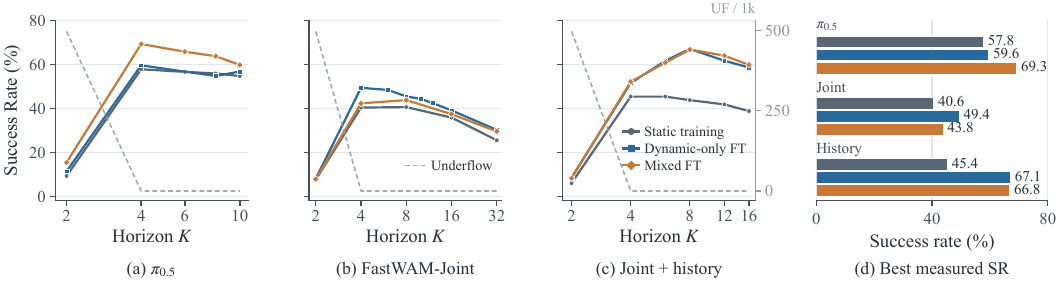}
\caption{Dynamic-data fine-tuning under asynchronous execution. \textbf{(a--c)} Measured $K$ sweeps for three models and steady underflow (right axis). \textbf{(d)} Best-measured SR for each model and training recipe.}
\label{fig:training_ksweep}
\end{figure*}

We compare $\pi_{0.5}$ (JAX), single-frame FastWAM-Joint, and FastWAM-Joint with five video frames and sixteen prior actions. Each uses static training, dynamic-only fine-tuning, and mixed dynamic--static fine-tuning. The Joint recipes run for 3{,}000 steps; $\pi_{0.5}$ dynamic-only and mixed recipes use 1{,}500 and 3{,}000 steps, so their difference does not isolate data composition. \Cref{fig:training_ksweep} reports measured even-$K$ sweeps under frozen latency; exact configurations and best-point metrics are in \cref{tab:training_config,tab:training_best}.

At their best measured horizons, $\pi_{0.5}$ reaches 57.8\% (static), 59.6\% (dynamic only), and 69.3\% (mixed); single-frame Joint reaches 40.6\%, 49.4\%, and 43.8\%; and history-frame Joint reaches 45.4\%, 67.1\%, and 66.8\%. Under the matched 3{,}000-step dynamic-only recipe, adding history changes the gain from 8.8 to 21.7 points and shifts the best measured horizon from $K{=}4$ to $K{=}8$. This associates temporal conditioning with greater use of the dynamic data in these variants, without establishing superiority for every horizon or architecture.

\finding{Dynamic fine-tuning gains vary with model design and training recipe, motivating architectures tailored to dynamic manipulation.}{%
Each model improves under at least one dynamic-data recipe, but the magnitude depends on model design and training configuration.
}

\subsection{Real-Robot Validation}
\label{sec:exp:realrobot}

We deploy $\pi_{0.5}$, FastWAM-Joint, and history-frame FastWAM-Joint on a Franka Research~3 at 15\,Hz using two external cameras and one wrist camera. Policies are trained on physical dynamic demonstrations; within each model, Async and RTC share a checkpoint. Hardware, sensing, and training details are in \cref{app:realrobot_details}.

Four conditions cover conveyor handover, whack-a-mole, and car pursuit on smooth-polynomial and zigzag paths. Each manipulation condition uses 50 trials; whack-a-mole uses 50 sessions of 30\,s. All six model--execution configurations are evaluated at $K\in\{4,8\}$. \Cref{fig:realrobot_tasks} shows the task sequences; exact values and complete sequences are in \cref{tab:realrobot,app:realrobot_details}.

\begingroup
\setlength{\intextsep}{4pt}
\begin{figure}[t]
\centering
\captionsetup{skip=3pt}
\captionsetup[subfigure]{skip=2pt}
\begin{subfigure}[t]{0.47\linewidth}
  \centering
  \includegraphics[width=\linewidth]{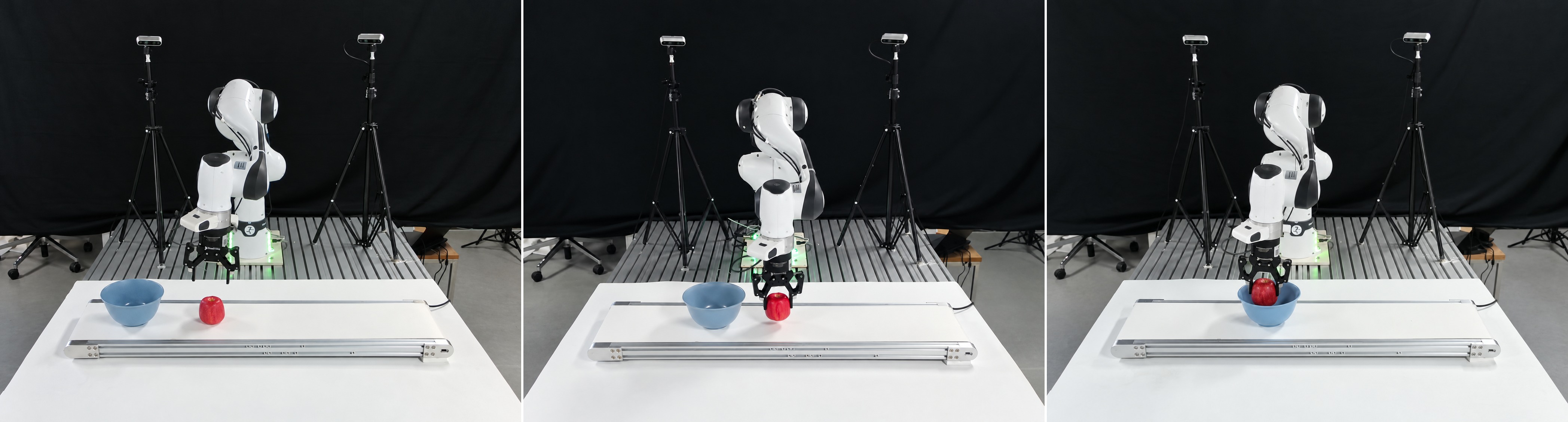}
  \caption{Conveyor handover}
\end{subfigure}\hfill
\begin{subfigure}[t]{0.47\linewidth}
  \centering
  \includegraphics[width=\linewidth]{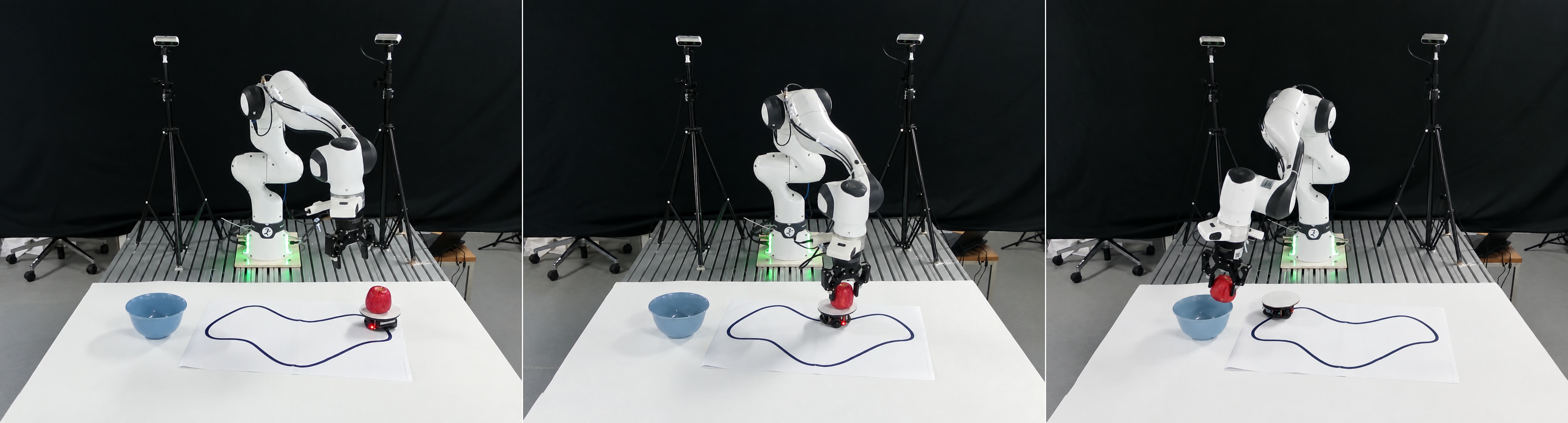}
  \caption{Smooth car pursuit}
\end{subfigure}

\vspace{1pt}
\begin{subfigure}[t]{0.47\linewidth}
  \centering
  \includegraphics[width=\linewidth]{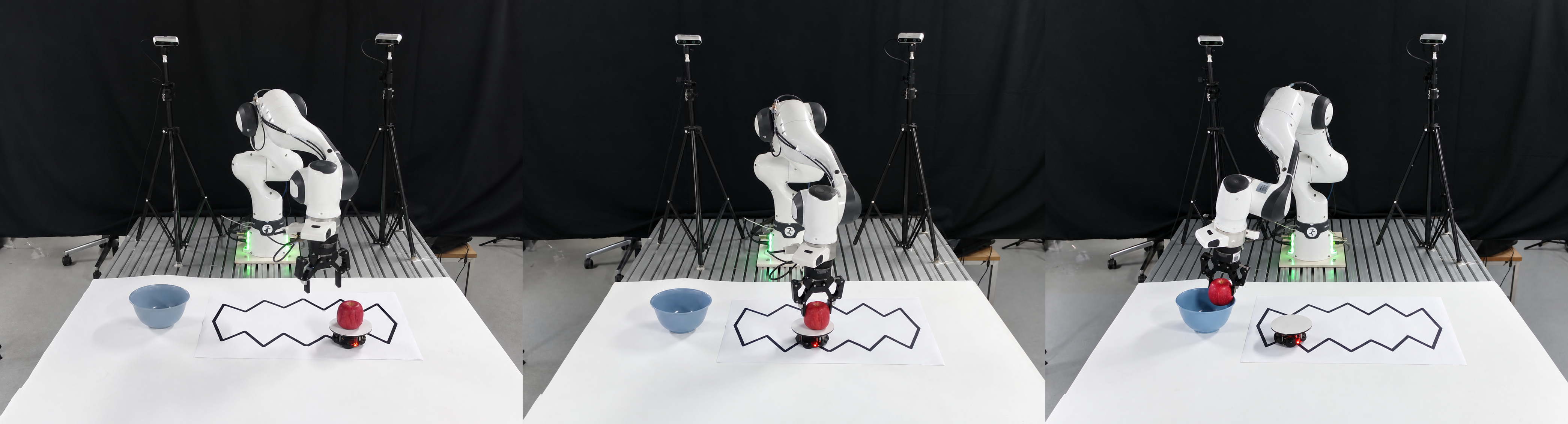}
  \caption{Zigzag car pursuit}
\end{subfigure}\hfill
\begin{subfigure}[t]{0.47\linewidth}
  \centering
  \includegraphics[width=\linewidth]{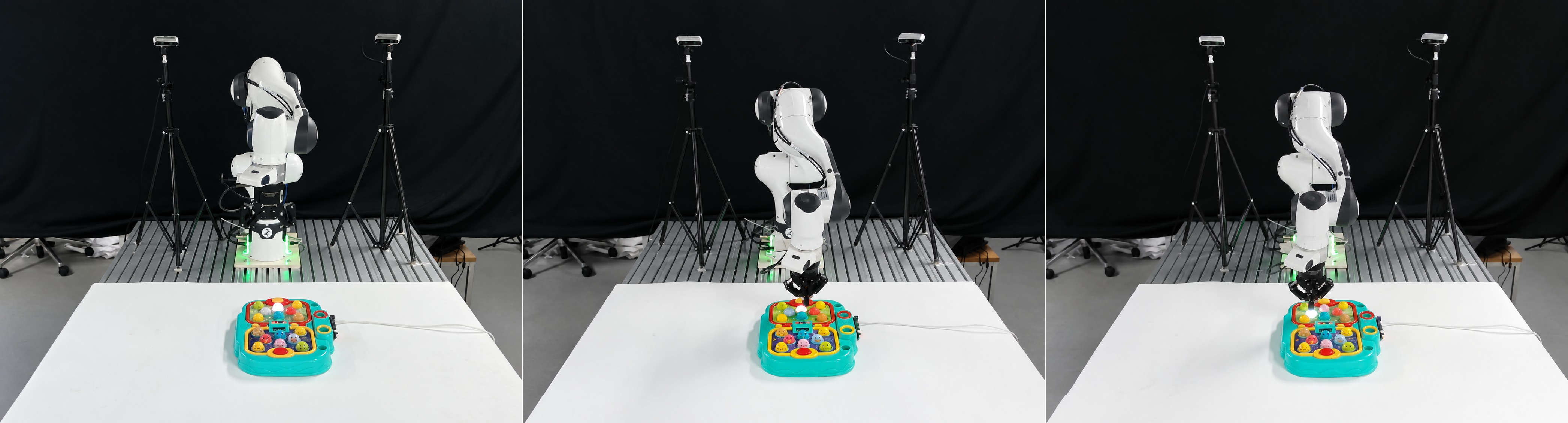}
  \caption{Whack-a-mole}
\end{subfigure}
\caption{Real-robot task sequences. Each panel shows three keyframes from one successful rollout; complete sequences are in the appendix.}
\label{fig:realrobot_tasks}
\end{figure}
\endgroup

The four tasks probe different temporal demands: conveyor handover and smooth pursuit require continuous interception, zigzag pursuit adds abrupt direction changes, and whack-a-mole requires repeated time-critical responses. We keep the model--execution pairs fixed and report both horizons for every task. At $K{=}4$, Async exceeds RTC on all three manipulation tasks for each model. Moving to $K{=}8$ lowers the aggregate for both $\pi_{0.5}$ settings and single-frame Joint Async, leaves single-frame Joint RTC unchanged, and improves both history-frame settings; whack-a-mole decreases for all six settings. \Cref{fig:realrobot_results} summarizes these results, showing model- and task-dependent horizon effects and providing a complementary physical check that execution timing can affect dynamic behavior.

\begingroup
\setlength{\intextsep}{4pt}
\begin{figure}[H]
\centering
\includegraphics[width=\linewidth]{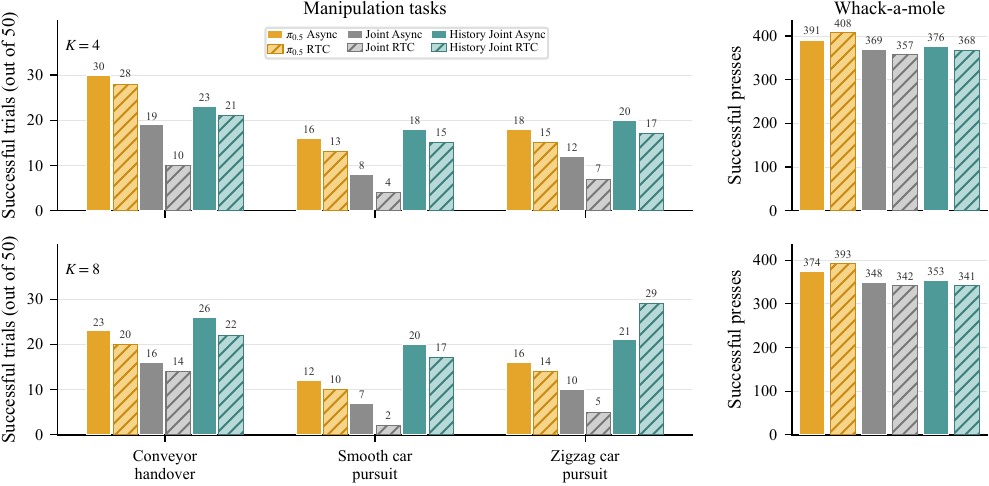}
\caption{Real-robot results at $K\in\{4,8\}$. The manipulation tasks report successful trials out of 50; whack-a-mole reports successful presses over $50\times30$\,s sessions.}
\label{fig:realrobot_results}
\end{figure}
\endgroup

\section{Limitations}
\label{sec:analysis}
\bench builds on \libero tasks and scenes, using prescribed trajectories and fixed calibrated latency to enable repeatable dynamic evaluation. It does not yet cover deployment-time latency fluctuations or more complex, unpredictable object motion. The fine-tuning study separates training and evaluation by motion phase while sharing tasks, paths, and speeds; it therefore primarily evaluates adaptation to specified dynamic conditions, with generalization to unseen tasks and motion conditions left for further study. The reported best action horizons are drawn from finite empirical sweeps, and their selection depends on the model, execution protocol, and evaluation conditions. Real-robot experiments provide complementary evidence for some timing-related findings, but cover a limited set of tasks and a single platform; the broader applicability of the simulation findings requires further validation.

\section{Conclusion}
\label{sec:conclusion}
We introduced \bench, incorporating controlled object motion and model-specific inference latency into all 40 \libero manipulation tasks to evaluate policies while the physical world continues to evolve. By replaying static demonstrations with online motion compensation, we further constructed a dataset of 9{,}000 dynamic demonstrations, providing a shared experimental basis for evaluation and training in dynamic manipulation.

Cross-model evaluation, execution-strategy comparisons, and dynamic fine-tuning experiments jointly show that dynamic manipulation performance depends on both the model and the timing of inference and execution. More frequent feedback must be coordinated with action-queue continuity, while suitable action horizons vary with the model and training configuration. All three evaluated model families improve under at least one dynamic-data recipe, but the gains differ, pointing to temporal information use and training recipes as important directions for model design in dynamic manipulation. Small-scale real-robot experiments also show that execution strategy and action horizon affect performance, providing complementary support for the corresponding simulation findings. \bench offers a reproducible platform for systematically studying these factors and their interactions.

\subsection*{Code and Data Availability}
Code, benchmark configurations, aggregate results, and supplementary videos are available from the
\href{https://renshaojie233.github.io/LIBERO-Dynamic/}{project page} and the
\href{https://github.com/renshaojie233/LIBERO-Dynamic}{public repository}.

\subsection*{Reproducibility Statement}The paper and supplement provide protocols, configurations, code, exact results, audits, and videos needed to reproduce the reported results with public checkpoints.

\subsection*{Ethics Statement}This work uses no human-subject or personal data. Robot experiments were supervised with safety controls; released artifacts contain no personal identifiers.

\subsection*{AI Use Statement}Generative AI assisted language editing, translation, and literature discovery and summarization. The authors verified all AI-assisted text and citations and take responsibility for the final paper. No experimental data were AI-generated.

\bibliographystyle{plainnat}
\bibliography{refs}
\clearpage
\appendix
\clearpage
\section{Appendix}
\label{app:main}

\subsection{Additional Benchmark Construction Details}\label{app:benchmark_details}
Among the 40 tasks, 25 contain two or three independently moving movable objects, five center on moving fixtures such as cabinet doors or stoves, and ten use one movable object; together they cover 23 target-object types. The linear family uses constant-speed piecewise-linear reciprocation through three waypoints. The Catmull--Rom family interpolates 3--8 waypoints into a $C^1$-continuous closed curve traversed at constant arc-length speed. The zigzag family follows 5--8 waypoints with unsmoothed intermediate turns.

All retained platform--target configurations remain in the static-friction regime at the evaluated speeds, including zigzag turns, and selective contact masks prevent the platform from interacting with non-target objects. Dynamic-demo generation begins object motion at control step 30 and caps episodes at 600 steps. The evaluation phase advances the scene by 40 steps before the first observation; the compact training split draws 450 episodes from each of the other four phase ranks.

\subsection{Metric Display Details}\label{app:metric_details}
References for $E_T$ and $E_P$ match successful records by $(s,i)$ and minimize step count and end-effector path separately. The cross-model and training tables (\cref{tab:crossmodel,tab:training_best,tab:ablation}) use a frozen five-base-policy reference: $\pi_{0.5}$, $\pi_0$, VLA-JEPA, OpenVLA-OFT, and FastWAM-Joint. Ratios are capped at 1, and a success without a matching successful reference receives ratio 1; scoring a row does not update the reference for other rows. The strategy table (\cref{tab:strategy}) instead uses per-condition minima across its five listed configurations, including conditions solved by only a subset. The two radar panels use a joint nine-model reference, so their efficiency values are not directly interchangeable with table values. Radar display ranges are fixed to SR 0--80\%, latency 40--120\,ms (reversed), $E_T/E_P$ 0.40--0.85, and $E_I$ 0--12 successes per 1{,}000 requests. Tables retain values before display mapping. Because the efficiency quantities also depend on successful outcomes, radar area is not aggregated into a scalar score.

\subsection{Dynamic-Demonstration Generation Details}\label{app:data_generation}
\paragraph{Static-reference selection and phase annotation.}
For each task, the generator scans the official \libero HDF5 demonstrations in a fixed order and replays each candidate with dynamic motion disabled. A candidate is valid only if the original BDDL goal succeeds. The primary reference minimizes successful completion step, then demonstration length and scan order, among the successful candidates found by the deterministic scan; the selected key is cached per task. A task-specific fallback may instead request another explicitly identified demonstration, which is replay-validated before use. Thus, reference changes occur only through recorded fallback specifications rather than random resampling.

The selected reference is executed once more to record the end-effector position, gripper command, and poses of task-relevant objects and goal receptacles at every control step. A gripper-close event near an object marks grasp onset, and the first later reopening marks release. For articulated fixtures, the closest end-effector approach anchors the interaction phase. These annotations determine when target tracking should affect the action and prevent compensation from being applied throughout unrelated parts of the demonstration.

Before formal generation and evaluation, each trajectory configuration is preflighted with the robot stationary. Configurations are excluded if a target leaves the safe workspace or camera view, or collides with non-target scene objects.

\paragraph{Online target-relative compensation.}
Let the reference end-effector position and target pose at step~$t$ be $p_t^{e}$ and $(p_t^{o},R_t^{o})$, and let the measured target pose during dynamic replay be $(\tilde p_t^{o},\tilde R_t^{o})$. The generator preserves the reference target--end-effector relation in the current target frame:
\begin{equation}
\hat p_t^{e}=\tilde p_t^{o}+\tilde R_t^{o}(R_t^{o})^{-1}(p_t^{e}-p_t^{o}),
\end{equation}
and uses $\hat p_t^{e}-p_t^{e}$ as a candidate displacement correction. Around grasp, the weight ramps up before contact, remains active briefly after grasp, and then fades; for placement, compensation is handed from the picked object to its destination; fixture compensation remains active through the recorded interaction. If no temporal annotation applies, candidates are weighted by spatial proximity. For multiple candidates, temporally active targets take precedence. The aggregate correction is clipped, exponentially smoothed, and converted into a bounded translational update to the reference action.

The primary route uses template replay with online compensation. Recorded task-specific fallbacks may use target-relative pick-and-place servoing, retain a static prefix or suffix, or adjust bounded grasp/place parameters. Every route advances the robot and physics only through standard \texttt{env.step(action)} calls. Object motion is induced through contact friction between the carrier and target; the generator never teleports the target or writes its pose to complete a demonstration.

\paragraph{Strict failure detection.}
The task predicate alone is insufficient for acceptance. The strict validator also requires a clean rollout status, zero assisted steps, no forbidden state edit, and positive motion evidence. For platform-carried targets, recorded platform displacement must reach 5\,mm; for other dynamic fixtures, either translation must reach 5\,mm or rotation must reach 0.03\,rad. The physical checks in \cref{tab:data_generation_checks} are evaluated before robot contact so that intended grasp-induced separation does not appear as a conveyor-following failure. Packaging additionally rejects missing episode arrays or inconsistent trajectory lengths.

\begin{table}[!ht]
\centering
\caption{Strict physical checks used for generated demonstrations. Crossing any rejection threshold prevents the episode from entering the released dataset.}
\label{tab:data_generation_checks}
\footnotesize
\setlength{\tabcolsep}{4pt}
\begin{tabular*}{\linewidth}{@{\extracolsep{\fill}}p{0.30\linewidth}p{0.62\linewidth}@{}}
\toprule
Check & Rejection condition \\
\midrule
Task and rollout & non-successful BDDL goal, runtime error, assisted step, or forbidden state edit \\
Motion evidence & platform/fixture motion below the thresholds above \\
Carrier path & maximum platform path error $>0.01$\,m \\
Target following & pre-contact object--carrier relative error $>0.08$\,m \\
Object stability & relative tilt $>35^{\circ}$ or downward displacement $>0.04$\,m \\
Visibility & target outside the diagnostic bird-view projection for more than 25\% of sampled pre-contact steps \\
\bottomrule
\end{tabular*}
\end{table}

\paragraph{Bounded retry and provenance.}
If the primary rollout fails strict validation, the generator traverses a finite, deterministic list of task-specific fallback specifications. A fallback may select another replay-validated static reference, switch between compensated-template and target-relative servo control, enable a grasp guard, or adjust fixed grasp/place margins. Scripted grasp routes may also try a finite center-and-neighborhood offset pattern; they do not keep sampling until success. Validation is rerun after every attempt and the cascade stops at the first strict success. If no candidate passes, the item remains failed, the process exits with an incomplete audit, and the episode is excluded from the packed HDF5 release. The manifest preserves the item identifier, source demonstration, motion condition and realized phase, controller route, per-attempt failure reasons, physical diagnostics, output paths, and random seed. The phase-partition auditor separately fails on count, uniqueness, coverage, or train/test-overlap violations.

\subsection{Representative Dynamic Demonstrations}
\label{app:task_gallery}
\Cref{fig:dynamic_task_gallery} selects one representative task from each \libero subset: placing the black bowl between the plate and ramekin onto the plate (Spatial), placing tomato sauce in the basket (Object), placing the bowl on the stove (Goal), and placing both moka pots on the stove (LIBERO-10). Each task is shown under linear ping-pong, curvilinear spline, and multi-waypoint zigzag motion at 0.09\,m/s. In every cell, the first three frames span the pre-contact approach and the fourth is the final successful state. The green curve is the episode-specific object path projected into AgentView using the stored MuJoCo camera, with arrows indicating motion direction. Selection is independent of the success or failure of any evaluated model. The electronic supplement provides the complete dual-view videos and eight-frame sequences for all 12 conditions.

\clearpage
\begin{center}
  \centering
  \vspace*{0.12in}
  \includegraphics[width=0.92\linewidth]{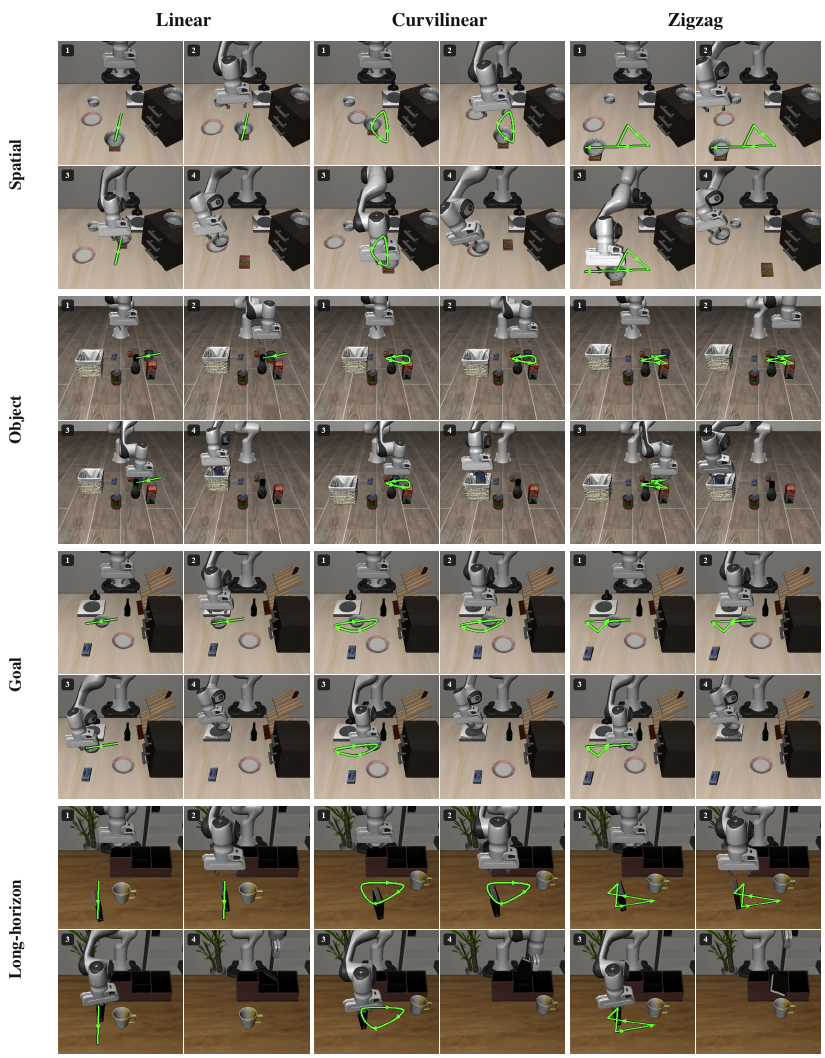}
  \captionof{figure}{Dynamic demonstrations for four representative tasks under the three motion families. In each cell, frames 1--3 span the measured approach from episode start to first robot--target contact, while frame 4 is the final successful state. The target position is read from every stored MuJoCo state and projected directly into AgentView; green arrows indicate direction. The narrow foreground trajectory remains legible after the figure is reduced to page width. Rows identify task subsets and columns identify motion families.}
  \label{fig:dynamic_task_gallery}
\end{center}
\clearpage

\subsection{Motion-Family Breakdown}
\label{app:motion_breakdown}
\Cref{tab:motion_breakdown} decomposes the cross-model result by trajectory family at the same asynchronous $K^*$ used in \cref{tab:crossmodel}. Each family contains 600 episodes (40 tasks $\times$ 5 trajectories $\times$ 3 speeds). We do not retune $K$ separately by family, so the rows preserve the deployment configuration used for the overall comparison. The breakdown is descriptive: the three families vary path geometry and do not define an ordered difficulty scale.

\begin{table}[!ht]
\centering
\caption{Asynchronous success rate by motion family at each model's overall best measured eligible horizon $K^*$. Each family contains 600 episodes; All contains 1{,}800.}
\label{tab:motion_breakdown}
\small
\setlength{\tabcolsep}{5pt}
\begin{tabular*}{\linewidth}{@{\extracolsep{\fill}}lccccc@{}}
\toprule
Model & $K^*$ & \Lonefam & \Ltwofam & \Lthreefam & All \\
\midrule
$\pi_{0.5}$ (JAX) & 4 & 62.7 & 51.2 & 59.7 & 57.8 \\
FastWAM-Joint & 8 & 45.2 & 36.3 & 40.3 & 40.6 \\
VLA-JEPA & 4 & 39.3 & 33.3 & 34.5 & 35.7 \\
FastWAM & 8 & 34.8 & 29.8 & 31.5 & 32.1 \\
$\pi_0$ (JAX) & 8 & 37.5 & 27.5 & 29.5 & 31.5 \\
OpenVLA-OFT & 4 & 30.8 & 25.0 & 26.0 & 27.3 \\
\bottomrule
\end{tabular*}
\end{table}

\subsection{Complete Cross-Model Action-Horizon Sweeps}
\begin{figure}[!ht]
  \centering
  \includegraphics[width=\linewidth]{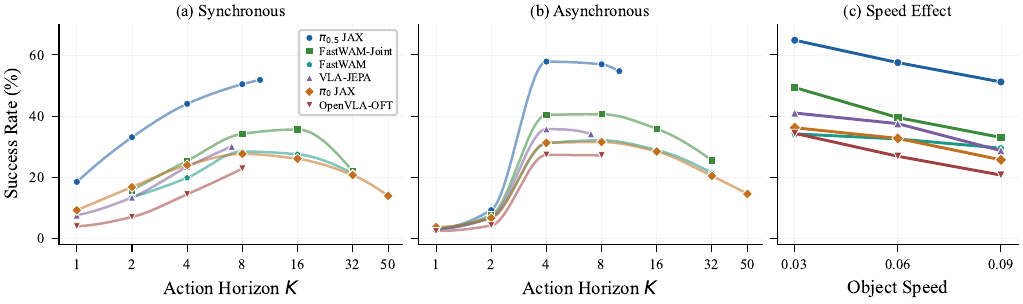}
  \caption{Cross-model action-horizon analysis on \bench: \textbf{(a)} synchronous and \textbf{(b)} asynchronous $K$ sweeps; \textbf{(c)} SR by object speed at each model's best measured asynchronous $K$.}
  \label{fig:ksweep}
\end{figure}

\Cref{fig:ksweep} reports the complete synchronous and asynchronous sweeps underlying the best-measured operating points in \cref{tab:crossmodel}, together with the per-speed asynchronous results at those operating points.

\subsection{Execution-Protocol Implementation Details}\label{app:execution_protocol}
Each policy produces a native action chunk of length~$H$, and evaluation retains the prefix specified by the effective action horizon~$K$. Measured latency~$\tau$ is quantized as $D=\lceil\tau f_c\rceil$ control steps at $f_c=20$~Hz; thus 60\,ms produces $D=2$ and a realized 100\,ms simulation delay. Synchronous hold submits a new observation after these~$K$ actions are consumed and locks the robot joints for the subsequent~$D$ delay steps; each cycle therefore contains~$K$ action steps and~$D$ waiting steps.

The asynchronous implementation maintains a first-in-first-out queue. When the remaining action count is at most $\lceil0.5K\rceil$, a background thread submits the next inference request while the robot continues consuming queued actions. When the new chunk arrives, overlapping positions are blended with weights $0.3$ for the old action and $0.7$ for the new action, and the non-overlapping suffix is appended. The implementation uses the producer--consumer structure of the LeRobot asynchronous queue~\citep{cadene2026lerobot}; the trigger ratio and blending weights are evaluation settings used in this work.

RTC also prefetches from a queue: the fixed setting triggers when at most 10 actions remain, while the proportional setting triggers at $\lceil\theta K\rceil$ remaining actions ($\theta=0.5$). No new request is submitted while one is in flight or awaiting delivery; the fixed threshold is not clipped to $K$. RTC additionally conditions the policy on unexecuted actions and the expected delay~$d$. The first~$d$ returned actions, corresponding to elapsed time, are removed, and the remainder replaces the current queue without weighted blending~\citep{black2025rtc}. VLASH submits a request when the queue length is at most $\min(K,d)$, using the current image and a causal expected-execution-time robot state predicted from committed actions; the predictor has no access to future scene state. The returned chunk replaces the queue intact after committed actions are exhausted, with neither prefix dropping nor weighted blending~\citep{vlash2025}.

Inference latency includes observation preprocessing, data transfer when applicable, model forward computation, and action postprocessing, ending when the complete action chunk reaches the controller. Each distinct model deployment path is calibrated using single-request inference on representative \libero observations; RTC calibration includes processing of its conditioning inputs. Warm-up requests are discarded. When multiple independent processes are used, their retained samples are pooled, and the median is frozen and converted to $d=\lceil\tau f_c\rceil$ control steps.

The $\pi_{0.5}$ JAX, FastWAM, $\pi_{0.5}$ PyTorch Sync/Async, and RTC calibrations each use one run with 8 warm-up and 88 retained requests. FastWAM-Joint and history-frame Joint each use one run with 20 warm-up and 100 retained requests. $\pi_0$, OpenVLA-OFT, and VLASH each use two independent processes with 8 warm-up and 88 retained requests per process; VLA-JEPA uses two processes with 30 warm-up and 300 retained requests per process. Fine-tuned checkpoints that preserve the deployment path reuse the corresponding base calibration. The electronic supplement's \texttt{tables/latency\_calibrations.csv} records every calibration identifier, exact frozen value, sample count, and evidence granularity.

Underflow accounting separates startup and steady state. An empty queue before the first queued prediction has been executed is startup underflow; an empty queue afterward is steady underflow. Figures report $1000N_{\mathrm{steady}}/N_{\mathrm{exec}}$, where $N_{\mathrm{exec}}$ is the total number of executed control steps in the episode.

\subsection{RTC Settings and Complete Strategy Sweeps}
\label{app:rtc}
Both RTC settings use a calibrated latency of 170\,ms. Fixed${=}$10 denotes a prefetch threshold of 10 queued actions, whereas $\theta{=}0.5$ denotes a threshold proportional to $K$; neither denotes denoising steps or inference delay.
The fixed${=}$10 setting peaks at 37.83\% and $\theta{=}0.5$ at 32.94\%, both at $K{=}16$.
We retain both settings here to make the main-text choice explicit; fixed${=}$10 does not outperform $\theta{=}0.5$ at every measured $K$.
\begin{figure}[!ht]
  \centering
  \includegraphics[width=\linewidth]{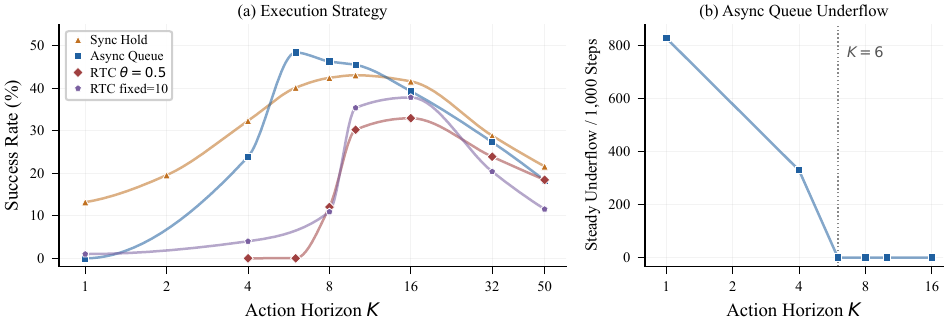}
  \caption{Extended $\pi_{0.5}$ (PyTorch) strategy sweeps, including two RTC settings and asynchronous underflow. Markers are measured results; smooth SR curves are visual guides.}
  \label{fig:rtc_settings}\end{figure}

\subsection{Best-Measured Execution-Strategy Results}
\label{app:strategy_results}
\Cref{tab:strategy} reports the per-speed success rates and efficiency metrics for the execution settings summarized by the main-text strategy sweep and radar plot. Each row uses that setting's best measured eligible action horizon.
\begin{table}[!ht]
\centering
\caption{Execution-strategy results on \bench at the best measured eligible horizon for each setting (1{,}800 episodes per row). Sync, Async, and RTC use $\pi_{0.5}$ (PyTorch); VLASH uses its official checkpoint. Zero-latency Sync is an idealized timing diagnostic.}
\label{tab:strategy}
\small
\setlength{\tabcolsep}{3.5pt}
\begin{tabular*}{\linewidth}{@{\extracolsep{\fill}}lcc cccc ccc@{}}
\toprule
 & $\tau$ & & \multicolumn{4}{c}{SR (\%)} & & & \\
\cmidrule(lr){4-7}
Strategy & (ms) & $D$ & 0.03 & 0.06 & 0.09 & All & $E_T$ & $E_P$ & $E_I$ \\
\midrule
VLASH & 104 & 3 & \textbf{65.7} & \textbf{54.7} & \textbf{46.0} & \textbf{55.4} & \textbf{.671} & \textbf{.689} & 5.61 \\
Sync Hold (zero-lat.) & 0 & 0 & 57.7 & 49.8 & 43.7 & 50.4 & .658 & .664 & \textbf{10.11} \\
Async Queue & 147 & 3 & 55.7 & 45.2 & 44.3 & 48.4 & .634 & .648 & 3.56 \\
Sync Hold & 147 & 3 & 48.7 & 42.5 & 38.0 & 43.1 & .593 & .618 & 9.85 \\
RTC (fixed${=}$10) & 170 & 4 & 44.7 & 38.3 & 30.5 & 37.8 & .556 & .565 & 5.02 \\
\bottomrule
\end{tabular*}
\end{table}

\subsection{Paired Execution-Strategy Behavior}
\label{app:strategy_behavior}
Table~\ref{tab:strategy_behavior} supplements the main five-axis metrics for the three calibrated $\pi_{0.5}$ configurations. Means are computed on 343 common successful $(s,i)$ pairs; all 1{,}029 matched traces are valid, with no exclusions. Extra successes contribute to the main SR and efficiency metrics but do not enter this matched subset.
Control steps exclude synchronous inference waits, and paths connect recorded EEF positions only. $\Delta a$ averages the recorded per-step action-change norm within each episode. $A_{\mathrm{EEF}}$ estimates acceleration using actual simulation timestamps, followed by equal episode weighting; uniformly spaced samples are not assumed. Unrecorded motion between synchronous samples remains a limitation. Smaller action changes or acceleration do not necessarily imply better dynamic tracking; these are behavioral descriptions, not a scalar ranking.
\begin{table}[!ht]
\centering
\caption{Behavioral metrics on 343 matched evaluation conditions successful under all compared strategies. $\Delta a$ measures action change, and $A_{\mathrm{EEF}}$ is a control-sampled acceleration estimate.}
\label{tab:strategy_behavior}
\small
\setlength{\tabcolsep}{3.5pt}
\begin{tabular*}{\linewidth}{@{\extracolsep{\fill}}lccccc@{}}
\toprule
Strategy & $K$ & Steps & Path (m) & $\Delta a$ & $A_{\mathrm{EEF}}$ (m/s$^2$) \\
\midrule
Async Queue & 6 & 167.1 & 1.159 & 0.142 & 0.379 \\
Sync Hold & 10 & 183.5 & 1.170 & 0.133 & 0.605 \\
RTC (fixed${=}$10) & 16 & 174.3 & 1.236 & 0.114 & 0.364 \\
\bottomrule
\end{tabular*}
\end{table}

\subsection{Dynamic Fine-Tuning Configuration}
\label{app:training_config}
All fine-tuning runs start from the corresponding static checkpoint: the official $\pi_{0.5}$ \libero checkpoint, the 60K-step single-frame Joint checkpoint, or the 60K-step history-frame Joint checkpoint. The dynamic set contains 1{,}800 demonstrations (284{,}502 frames at 20\,Hz) drawn from training phases whose episode IDs do not appear in evaluation; tasks, paths, and speeds remain condition-matched. Mixed fine-tuning uses the union of these demonstrations and 1{,}693 repaired static \libero demonstrations (273{,}465 frames); dynamic episodes therefore constitute 51.5\% of the union by episode count. For the two Joint models, the dynamic gripper signal is mapped to the same $[0,1]$ convention as the static data. For $\pi_{0.5}$, the static data are converted to the model's native gripper convention and LeRobot v2.1 format. Normalization statistics are recomputed for each fine-tuning dataset.

\begin{table}[H]
\centering
\caption{Dynamic fine-tuning configurations. Batch is the effective global batch. The mixed set is the union of the dynamic and repaired static demonstrations described above.}
\label{tab:training_config}
\footnotesize
\setlength{\tabcolsep}{2.5pt}
\begin{tabular*}{\linewidth}{@{\extracolsep{\fill}}llcccc@{}}
\toprule
Model & FT data & Steps & Batch & Peak LR & Warm-up \\
\midrule
\multirow{2}{*}{$\pi_{0.5}$} & Dynamic only & 1{,}500 & 256 & $5{\times}10^{-5}$ & 75 \\
 & Dynamic + static & 3{,}000 & 256 & $5{\times}10^{-5}$ & 150 \\
\midrule
\multirow{2}{*}{Joint} & Dynamic only & 3{,}000 & 128 & $1{\times}10^{-4}$ & 150 \\
 & Dynamic + static & 3{,}000 & 128 & $1{\times}10^{-4}$ & 150 \\
\midrule
\multirow{2}{*}{History Joint} & Dynamic only & 3{,}000 & 128 & $1{\times}10^{-4}$ & 150 \\
 & Dynamic + static & 3{,}000 & 128 & $1{\times}10^{-4}$ & 150 \\
\bottomrule
\end{tabular*}
\end{table}

The two Joint variants use AdamW with $\beta=(0.9,0.95)$, weight decay $10^{-2}$, gradient-norm clipping at 1.0, bf16 precision, and a cosine learning-rate schedule. Single-frame Joint receives one current observation and predicts 32 actions. History-frame Joint receives five consecutive observations and the preceding 16 executed actions; it predicts a 32-step sequence whose first 16 actions form the clean conditioning prefix, and the loss is applied to the 16 future actions. Dynamic-only Joint uses ZeRO-1 on eight GPUs; history-frame dynamic-only and mixed fine-tuning use DDP on eight and four GPUs, respectively. $\pi_{0.5}$ uses FSDP on eight GPUs and follows the OpenPI fine-tuning implementation with gradient clipping at 1.0, no EMA, and cosine decay to zero. Because the $\pi_{0.5}$ dynamic-only and mixed recipes differ in both data composition and training steps, their comparison is reported descriptively rather than as an isolated data-mixture ablation.
\clearpage
\subsection{Best-Measured Training Results}
\label{app:training_best}
The following table retains the complete per-speed success rates and efficiency metrics underlying Figure~\ref{fig:training_ksweep}(d). Each row uses the best result among its measured action horizons; the fixed-$K{=}8$ controlled comparison follows in the next section.
\begin{table}[H]
\centering
\caption{Results for three training recipes under asynchronous execution. Each row uses its best measured horizon $K^*$ and 1{,}800 episodes.}
\label{tab:training_best}
\small
\setlength{\tabcolsep}{1.5pt}
\begin{tabular*}{\linewidth}{@{\extracolsep{\fill}}llccc cccc ccc@{}}
\toprule
 & & $\tau$ & & & \multicolumn{4}{c}{SR (\%)} & & & \\
\cmidrule(lr){6-9}
Model & Training data & (ms) & $D$ & $K^*$ & 0.03 & 0.06 & 0.09 & All & $E_T$ & $E_P$ & $E_I$ \\
\midrule
\multirow{3}{*}{$\pi_{0.5}$ (JAX)} & Static & \multirow{3}{*}{60} & \multirow{3}{*}{2} & 4 & 64.8 & 57.5 & 51.2 & 57.8 & .715 & .727 & 3.14 \\
 & Dynamic only & & & 4 & 70.5 & 63.3 & 45.0 & 59.6 & .726 & .719 & 3.38 \\
 & Dynamic + static & & & 4 & 81.5 & 70.8 & 55.7 & 69.3 & .788 & .779 & 4.50 \\
\midrule
\multirow{3}{*}{FastWAM-Joint} & Static & \multirow{3}{*}{64} & \multirow{3}{*}{2} & 8 & 49.3 & 39.5 & 33.0 & 40.6 & .593 & .606 & 3.70 \\
 & Dynamic only & & & 4 & 64.7 & 48.5 & 35.0 & 49.4 & .651 & .671 & 2.51 \\
 & Dynamic + static & & & 8 & 52.8 & 43.5 & 35.1 & 43.8 & .603 & .624 & 3.27 \\
\midrule
\multirow{3}{*}{\shortstack[l]{FastWAM-Joint\\+ history}} & Static & \multirow{3}{*}{89} & \multirow{3}{*}{2} & 4 & 56.8 & 43.0 & 36.3 & 45.4 & .610 & .622 & 2.15 \\
 & Dynamic only & & & 8 & 76.2 & 70.3 & 54.8 & 67.1 & .784 & .773 & 8.40 \\
 & Dynamic + static & & & 8 & 77.8 & 70.8 & 51.8 & 66.8 & .786 & .777 & 8.45 \\
\bottomrule
\end{tabular*}
\end{table}

\subsection{Dynamic Fine-Tuning at a Fixed Action Horizon}
\label{app:training_ablation}
The fixed-$K{=}8$ comparison separates training changes from action-horizon tuning. Table~\ref{tab:ablation} additionally compares zero-latency Sync and calibrated-latency Async for one selected fine-tuning recipe per architecture; the main text reports all available recipes under Async.
Zero-latency synchronous results are diagnostic controls under ideal feedback and are excluded from the main results. Their difference from calibrated asynchronous execution changes both timing and execution strategy, so it cannot isolate the effect of latency alone.
\begin{table}[!ht]
\centering
\caption{Fixed-$K$ comparison before and after dynamic-data fine-tuning on \bench ($K{=}8$; 1{,}800 episodes per configuration). Sync is zero-latency, and Async uses calibrated latency.}
\label{tab:ablation}
\small
\setlength{\tabcolsep}{5pt}
\begin{tabular*}{\linewidth}{@{\extracolsep{\fill}}llcc@{}}
\toprule
Model & Training & Sync (\%) & Async (\%) \\
\midrule
\multirow{2}{*}{Single-frame Joint}
  & Static base         & 43.0 & 40.6 \\
  & Dynamic-only FT     & \textbf{52.0} & \textbf{45.5} \\
\midrule
\multirow{2}{*}{History Joint}
  & Static base         & 43.5 & 43.8 \\
  & Dynamic-only FT     & \textbf{62.4} & \textbf{67.1} \\
\midrule
\multirow{2}{*}{$\pi_{0.5}$}
  & Official (static)   & 59.3 & 55.9 \\
  & Mixed FT            & \textbf{65.4} & \textbf{63.8} \\
\bottomrule
\end{tabular*}
\end{table}

\clearpage

\subsection{Real-Robot Task Details and Full Rollouts}
\label{app:realrobot_details}
The real-robot study uses four dynamic evaluation conditions under the platform and observation setup described in \cref{sec:exp:realrobot}: conveyor handover, whack-a-mole, and car pursuit on smooth-polynomial and zigzag paths.
The two car trajectories are evaluated and presented separately because one changes direction continuously whereas the other contains abrupt turns.
The main-text zigzag panel uses frames 2, 4, and 6; the figures below retain all seven selected frames for every condition.

\paragraph{Hardware and sensing.}
The high-level policy runs at 15\,Hz on an RTX~5090 workstation, while an independent NUC executes the Franka controller at 1\,kHz. The two external cameras and the wrist camera stream at 30\,Hz into shared-memory buffers. At every 15\,Hz policy tick, frames are selected by nearest timestamp without image interpolation. The single-frame Joint input resizes each view to $224{\times}128$ and stacks the three views vertically into a $384{\times}224$ image; the history model uses the same spatial layout at the most recent five policy ticks. The robot state and action are both 8D: seven joint coordinates or native joint commands plus one absolute gripper command. Joint states and actions are z-score normalized, and the history model additionally conditions on the most recent 16 commands actually sent to the controller.

\paragraph{Training and deployment.}
The real-robot policies are trained on demonstrations from seven tasks collected at 15\,Hz. The $\pi_{0.5}$ and single-frame Joint checkpoints are trained for 20K steps. The history-frame checkpoint is initialized from the 20K-step Joint weights and trained for 30K additional steps with a five-frame visual history and a 16-action clean prefix. Within each architecture, Async and RTC use the same checkpoint, so their comparison changes the inference and execution protocol rather than the learned weights. Tables~\ref{tab:realrobot} and~\ref{tab:realrobot_config} report the exact outcomes and corresponding deployment settings.

\begin{table}[H]
\centering
\caption{Real-robot results for six model--execution configurations at $K{\in}\{4,8\}$. Conveyor and the two car-pursuit conditions report successful trials out of 50; whack-a-mole reports total successful presses across $50{\times}30$\,s sessions.}
\label{tab:realrobot}
\footnotesize
\setlength{\tabcolsep}{1.8pt}
\begin{tabular*}{\linewidth}{@{\extracolsep{\fill}}lllcccc@{}}
\toprule
Model & Exec. & $K$ & Conv. /50 & Smooth /50 & Zigzag /50 & Whack presses \\
\midrule
\multirow{4}{*}{$\pi_{0.5}$} & \multirow{2}{*}{Async} & 4 & 30 & 16 & 18 & 391 \\
 & & 8 & 23 & 12 & 16 & 374 \\
 & \multirow{2}{*}{RTC} & 4 & 28 & 13 & 15 & 408 \\
 & & 8 & 20 & 10 & 14 & 393 \\
\midrule
\multirow{4}{*}{FastWAM-Joint} & \multirow{2}{*}{Async} & 4 & 19 & 8 & 12 & 369 \\
 & & 8 & 16 & 7 & 10 & 348 \\
 & \multirow{2}{*}{RTC} & 4 & 10 & 4 & 7 & 357 \\
 & & 8 & 14 & 2 & 5 & 342 \\
\midrule
\multirow{4}{*}{\shortstack[l]{FastWAM-Joint\\[-3pt]+ history}} & \multirow{2}{*}{Async} & 4 & 23 & 18 & 20 & 376 \\
 & & 8 & 26 & 20 & 21 & 353 \\
 & \multirow{2}{*}{RTC} & 4 & 21 & 15 & 17 & 368 \\
 & & 8 & 22 & 17 & 29 & 341 \\
\bottomrule
\end{tabular*}
\end{table}

\begin{table}[H]
\centering
\caption{Deployment configuration for the real-robot results in Table~\ref{tab:realrobot}. All six settings run at 15\,Hz and are evaluated at $K{\in}\{4,8\}$. ``History'' denotes the five-frame input and 16-action clean prefix.}
\label{tab:realrobot_config}
\footnotesize
\setlength{\tabcolsep}{2.5pt}
\begin{tabular*}{\linewidth}{@{\extracolsep{\fill}}llcccc@{}}
\toprule
Model & Execution & Input & $H$ & Denoising & Evaluated $K$ \\
\midrule
\multirow{2}{*}{$\pi_{0.5}$} & Async & Current & 16 & 10 & 4, 8 \\
 & RTC & Current + RTC & 16 & 10 & 4, 8 \\
\midrule
\multirow{2}{*}{Joint} & Async & Current & 32 & 1 & 4, 8 \\
 & RTC & Current + RTC & 32 & 10 & 4, 8 \\
\midrule
\multirow{2}{*}{History Joint} & Async & 5 frames + 16 actions & 16$^{\dagger}$ & 1 & 4, 8 \\
 & RTC & History + RTC & 16$^{\dagger}$ & 10 & 4, 8 \\
\bottomrule
\end{tabular*}
\vspace{2pt}
\begin{minipage}{0.98\linewidth}
\footnotesize $^{\dagger}$The history model returns 32 actions in total; the first 16 are the clean conditioning prefix and only the future slice is eligible for execution.
\end{minipage}
\end{table}

Async inference runs in the background while the current plan is executed. RTC conditions denoising on the unexecuted action prefix and the expected inference delay, and the controller selects the newest time-aligned plan. For all methods, late or expired commands are skipped rather than replayed in a burst. If the valid queue is exhausted, the controller sends zero joint velocity and holds the last gripper command. Native joint commands are clipped by the same deployment safety envelope before conversion to 15\,Hz joint-velocity commands; the gripper remains an absolute command with 0 denoting closed and 1 denoting open.

\begin{figure}[!ht]
\centering
\includegraphics[width=\linewidth]{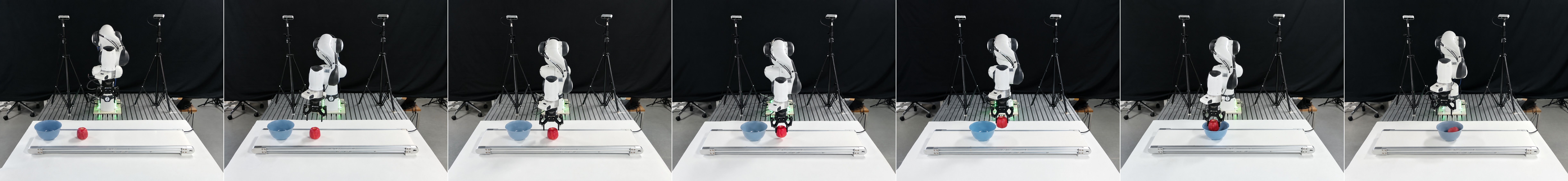}
\caption{Conveyor handover: seven keyframes from one successful episode.}
\label{fig:realrobot_conveyor_full}
\end{figure}

\begin{figure}[!ht]
\centering
\includegraphics[width=\linewidth]{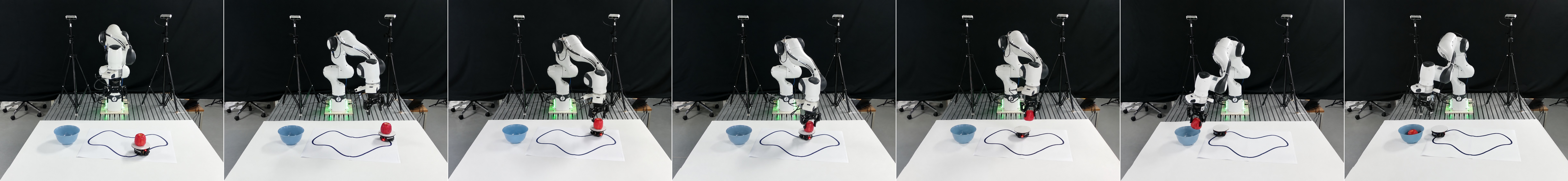}
\caption{Car pursuit with a smooth-polynomial trajectory: seven keyframes from one successful episode.}
\label{fig:realrobot_cart_smooth_full}
\end{figure}

\begin{figure}[!ht]
\centering
\includegraphics[width=\linewidth]{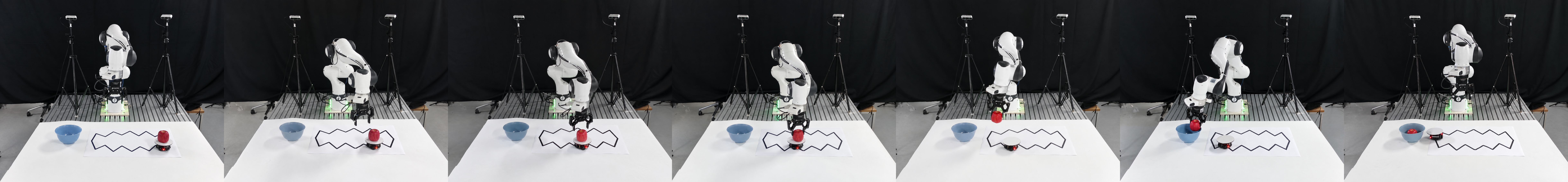}
\caption{Car pursuit with a zigzag trajectory: seven keyframes from one successful episode.}
\label{fig:realrobot_cart_zigzag_full}
\end{figure}

\begin{figure}[!ht]
\centering
\includegraphics[width=\linewidth]{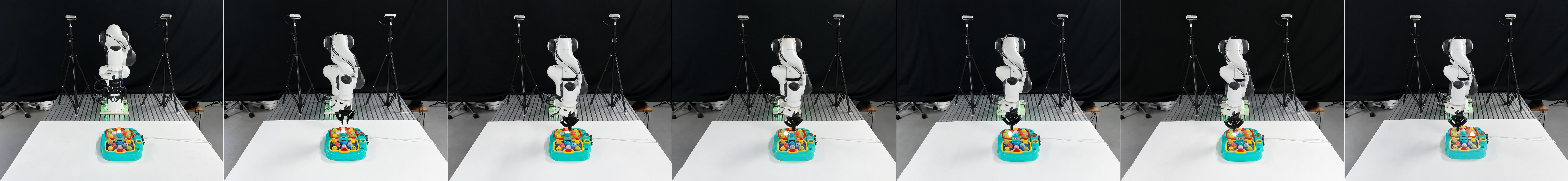}
\caption{Whack-a-mole: seven keyframes from one successful episode.}
\label{fig:realrobot_whack_full}
\end{figure}

\end{document}